# A lossless data hiding scheme in JPEG images with segment coding

Mingming ZHANG[a,*], Quan ZHOU[a], Yanlang HU[a]

[a]National Key Laboratory of Science and Technology on Space Microwave, China Academy of Space Technology, Xi'an 710100, China

**Abstract:**

In this paper, we propose a lossless data hiding scheme in JPEG images. After quantified DCT transform, coefficients have characteristics that distribution in high frequencies is relatively sparse and absolute values are small. To improve encoding efficiency, we put forward an encoding algorithm that searches for a high frequency as terminate point and recode the coefficients above, so spare space is reserved to embed secret data and appended data with no file expansion. Receiver can obtain terminate point through data analysis, extract additional data and recover original JPEG images lossless. Experimental results show that the proposed method has a larger capacity than state-of-the-art works.

## 1. Introduction

Data hiding is a technique that secret data is embedded into cover media which can be used in many applications, but it exists a drawback that there is an unnoticeable distortion to cover media since irreversible operations. Reversible data hiding (RDH) [1] can recover host images exactly after the extraction of secret data. In fields of law enforcement, medical systems and military imagery, etc, the technique is widely applied. RDH for images can be divided into three domains [2]: space domain [3]-[5], transform domain [6][7] and compressed domain [8][11].

Similarly according to technique realization, many RDH methods can be classified into five categories: lossless compression [12], difference expansion (DE) [13], histogram shifting (HS) [14], prediction-error expansion (PEE) [15] and integer-to-integer transform [16]. With these methods, good performance is achieved in both embedding capacity and visual quality, but these techniques are mostly applied in uncompressed images while majority of images are compressed as JPEG[17][18]. RDH approaches for JPEG images are divided into four types: quantizing DCT coefficients [19]-[24], modifying Quantization Table [25], modifying Huffman Table [26]-[30] and concealment in encrypted JPEG bitstreams [31][32]. While most of them still make a distortion to host images and can not lossless recover them. Lossless data hiding (LDH) can recover host images lossless after the extraction of secret data. Although some works can restore host images lossless, file expansion might occur and increased amount is surely larger than secret size.

To achieve LDH for JPEG images with no file expansion, Mobasseri [26] first proposed the algorithm to flip appended bits or a bit of variable-length-code (VLC) and applied VLC mapping technique, but it might result in decoding failure. Qian and Zhang [28] proposed the mapping algorithm to relate used VLCs to unused VLCs, and modified the Huffman Table in JPEG header.

Hu [29] improved the algorithm of Qian and Zhang [28] to calculate occurrence frequencies in used VLCs and map VLCs according to frequencies, thus further improved hiding capacity. Qiu [30] constructed an optimal mapping between used and unused VLCs in each category to achieve a high embedding rate, while it was quite complex. Chang [33] proposed a scheme that in framework of reserving-room-before-encryption by Ma [34], a part of JPEG bitstreams was compressed and encrypted to leave spare space to embed secret data. Then extraction of secret data and recovery of host images were both independent, but encryption had modified distribution of original data and could not have theoretical hiding capacity.

Although above LDH methods can preserve file size, hiding capacity is still insufficient. We propose a new LDH approach in JPEG bitstreams, and search for a high frequency as terminate point that coefficients below are coded in JPEG codes, while coefficients above are coded in the proposed Rxy codes. This segment coding can improve encoding efficiency that leaves spare space to embed secret data in file header and receiver can extract secret data and recover host images independently. We have achieved three goals: (1) file size does not increase; (2) both host images and secret data are lossless; (3) hiding capacity has greatly improved.

The rest of this paper is organized as follows: We first review previous works in Section 2. Then in Section 3, principle of segment coding and steps of hiding, extraction and recovery are introduced. Section 4 shows and explains results. Finally we make a conclusion in Section 5.

## 2. Previous works

## 2.1. VLCs mapping

In a 8×8 quantified DCT block, a non-zero coefficient is encoded in a format which contains variable-length-coefficients (VLC) and variable-length-integers (VLI). Each VLC is defined as (run, size) where run denotes zeroes number before and size denotes non-zero binary size, and VLI is corresponding non-zero binary code. There are 162 VLCs in Huffman coding table but only a part are used. According to bits size, 162 VLCs are classified into 16 categories $\{C_1, C_2, \ldots, C_{16}\}$.

To explore potential space, Qiu[30] constructed a LDH alternative embedding method using code mapping and reordering by analyzing relationships between used and unused VLCs. In VLC remapping, an unused VLC could be arbitrarily mapped to a specific used VLC, producing different combinations of unused VLCs mapping.

To avoid embedding capacity insufficient in individual category $C_i$, Qiu[30] also shuffled the orders of VLCs in $C_i$ that it was still $i$ bits, then a new Huffman tree was reconstructed. For examples in category $C_{16}$, 125 VLCs were included that embedding capacity was $\lfloor \log_2 125! \rfloor$ i.e. 695 bits. Then hiding capacity in category $C_i$ was equal to $EC_i = \text{maximize}\{EC_{i,a}, EC_{i,b}\}$, $EC_{i,a}$ was hiding capacity in improved mapping and $EC_{i,b}$ was in reordering.

## 2.2. Data hiding in encrypted JPEG bitstreams

Chang [33] constructed a separable RDH scheme for encrypted JPEG bitstreams based on structure by Ma [34]. He first compressed bitstreams with appended values [-3,-2,2,3] lossless to reserve space, then encrypted modified bitstreams. With the data embedding key, he hide secret data in least significant bits (LSB) of reserved space, and also encrypted quantization table that receiver could display images which were entirely different with original JPEG images without decryption key. Receiver could extract secret data with data embedding key then decrypt and decompress bitstreams to restore marked JPEG images which were lossless with original JPEG images.

# 3. Proposed methods

## 3.1. General framework

We decode bitstreams to quantified DCT blocks, search for the highest frequency with which absolute coefficient is larger than 2 in each block, and make a statistics on the whole frequencies and obtain the optimal high frequency as terminate point. Then segment coding scheme is proposed that coefficients below are encoded with JPEG codes, while coefficients above are encoded with the proposed Rxy codes, which results in bitstreams compression. Finally secret data is embedded in file header as comment to achieve lossless data hiding that file size does not increase.

### 3.1.1. Segment encode

After quantified DCT transform, data is encoded in the form of RSV (Run/Size and Value) and is concentrated in low frequencies while sparse in high frequencies. There are 162 alternating current (AC) Huffman codes in Table 1 [17], and in *Run/Size* column *Run* refers to zeroes number before non-zero coefficient, and *Size* refers to non-zero binary size, then appended bits refer to non-zero binary code. In theory, the whole 162 AC Huffman codes can be applied, but actually only a part are used especially in high frequencies.

In distribution of coefficients with high frequencies, there are often some zeroes before non-zero, and non-zero is quite small generally. *Run/Size* is joint coding of zeroes number before and non-zero binary size. If non-zero absolute is smaller than a threshold, no extra bits are needed then encoding length declines and encoding efficiency improves. In high frequencies, non-zero coefficients are so small that terminate point can be obtained above which absolutes are smaller than or equal to a threshold. In frequencies above, non-zero coefficients are encoded in the proposed Rxy codes, while in frequencies below, they are encoded in the standard codes.

[insert Table 1]

Where *code length* refers to zeroes number, *code word* refers to joint codes, *appended length* refers to non-zero binary size, *sum length* refers to joint codes size, *rate* refers to ratio of non-zero in joint codes. Suppose that *code length* is $m_1$ bits, *sum length* is $m$ bits, and *appended length* $m_2$ is equal to $m-m_1$ bits. Then *rate* can be as follows,

$$rate = m_2/m$$
$$m_2 = m - m_1$$
(1)

The Rxy encoding scheme is proposed that non-zeroes are in {-2,-1,1,2}, and encoding framework is shown in Fig.1. The former *Run* refers to zeroes number before with three bits, and following *xy* refers to non-zero.

[insert Figure 1]

*Run* encoding is shown in Table 2,

[insert Table 2]

'000' refers to encoding finished in a block.

*xy* refers to non-zero in {-2,-1,1,2} with 2 bits, and shown in Table 3.

[insert Table 3]

Joint encoding length is shorten result from reducing information entropy of non-zero. When zeroes number *M* is larger than 6, it is similar to the standard scheme that parameter *m* refers to quotient divided by *M* with 6, and parameter *n* refers to modular of *M* with 6.

$$m=\lfloor M/6 \rfloor$$
$$n=\mod(M,6) \quad (2)$$
$$M = m \times 6+n$$

When *M* is in {0,1,2,3,4,5,6}, joint encoding is shown in Table 4,

[insert Table 4]

When *M* is larger than 5, just $\lfloor M/6 \rfloor$ groups of '111' are added ahead.

The following is a custom table for the Rxy codes, as shown in Table 5,

[insert Table 5]

Compared with the standard Huffman table, *rate* of the proposed scheme is larger than the standard scheme except in *Run/Size* 0/2. The larger *rate* is, the higher proportion of non-zero is. In comparison of *code length*, the proposed scheme is also perfect. *code length* is almost smaller than the standard scheme except in *Run/Size* 0/1 and 0/2, which confirms that the proposed scheme has an advantage in sparse distribution of coefficients. When non-zero is in {-2,-1,1,2}, the more zeroes number is, the shorter *code length* is. At the same time, '000' which refers to finish is less than '1010' in the standard scheme.

The standard codes prefer to dense distribution of coefficients that absolute values are large in low frequencies, while in high frequencies coding efficiency is not perfect that absolute values are small and distribution is sparse that many bits are needed. When non-zero binary size is given, it is just needed to encode zeroes number. As long as zeroes number is not equal to 0, *code length* reduces greatly in the proposed scheme. After segment coding, spare space is reserved to embed secret data in header. In receiver, if terminate point is obtained, each 8×8 block can be separated, and secret data will be extracted. Since coefficients with different frequencies are not modified, JPEG images can be displayed without distortion. Specific embedding procedure is shown in Fig.2,

[insert Figure 2]

In this paper, the most important is to determine terminate point in segment coding. Since absolute value in the Rxy codes is smaller than or equal to 2, host bitstreams is decompressed into DCT coefficients, then make a statistics on frequencies with which absolute values are larger than 2. The maximum frequency in the whole blocks is terminate point.

$$\max j, st : |x_{i,j}| > 2, i \in [1,4096], j \in [1,64] \quad (3)$$

## 3.1.2. An example for Rxy codes

To have a better comprehension, we consider a 8×8 DCT block to illustrate our Rxy codes. First of all, we assume a block where the whole 64 coefficients are shown in Table 6.

[insert Table 6]

Corresponding *Run/Size* values are listed in Table 7,

[insert Table 7]

The maximum frequency with absolute value larger than 2 is the 12th, with which coefficient is 4 and Run/Size is 2/3. Above the point, coefficients are needed to recode in Rxy codes as listed:

For coefficients 0,0,0,-2, original binary codes are 111110111,10, then the coefficients are recoded as 100,00.

For coefficients 0,0,0,0,1, original binary codes are 111011,0, then the coefficients are recoded as 101,10.

For coefficients 0,0,0,0,0,0,0,0,-1, original binary codes are 111111001,1, then the coefficients are recoded as 111100,01.

As a finished marker, 1,0,1,0 are in JPEG codes while 0,0,0 are in Rxy codes.

Below the point, coefficients are coded in JPEG codes.

Above the point, original binary codes are 111110111,10,111011,0,111111001,1,1010 with 33 bits. While Rxy binary codes are 100,00,101,10,111100,01,000 with only 21 bits, then 12 bits are reserved to embed secret data.

### 3.1.3. Special cases for Rxy codes

In a 512×512 gray image, there exist some cases that no terminate point is found and Rxy codes fails. Then it is needed to record these special blocks and encode indexes as 16 bits. There are 64×64 i.e. 4096 blocks, then $\log_2(64 \times 64)$ i.e. 12 bits can confirm special block index. For example, the 3412th block is not suitable for Rxy codes, and 3412 is encoded as 110101010100. With 4 bits 0000 padded, 0000,110101010100 can be converted to 2 bytes with 0D54. Each special block index is encoded in 2 bytes and taken as appended data in JPEG header.

### 3.1.4. Additional data hiding

We decompress a 512×512 JPEG image to 4096 quantified 8×8 DCT blocks. In each block we find terminate point in which absolute value is larger than 2, and make a statistics of points to take the max value as terminate point in JPEG image.

In special blocks which are not suitable for Rxy codes, we need to record these block indexes separately. We assume that special block number is $m_N$, then each block index is encoded as 2 bytes and there are $m_N \times 2$ bytes corresponding to blocks $m_1, m_2, \ldots m_N$. For example, the 203th, 1268th, 3052th, 4025th blocks are not satisfied with the Rxy codes, we encode the indexes and take them as appended data.

A marker COM (comment) FFFE is used to insert in file header, and additional data is inserted following FFFE which are included with secret data and appended data. There are 64×64 DCT blocks in a 512×512 gray image, and 12 bits are needed to illustrate special blocks numbers, then we take 2 bytes as appended data part. There are 64 frequencies in each DCT block, then 1 byte is needed to represent terminate point. Secret size with 4 bytes is also needed as comment length in JPEG standard codes.

The constitute of additional data is shown in Fig.3,

[insert Figure 3]

In Fig.3, the first part is appended data, then the second part is secret data. The total embedded capacity is $M$ bytes, then secret data length $m_p$ is $M - m_N \times 2 - 9$ bytes.

## 3.2. Hide secret data

In conclusion of the schemes above, following steps are taken to hide secret data in this paper:

1. Decompress original JPEG image to DCT coefficients, and establish corresponding relationship between frequencies and codestreams.

2. Take the maximum frequency above which absolute value is larger than or equal to 2 as terminate point in each block.

3. Take the maximum of terminate points in each block as terminate point in image and record special block indexes.

4. In each block, encode coefficients below terminate point in the standard codes, while encode coefficients above in the proposed Rxy codes.

5. Calculate recompressed file size, and compare it with original JPEG file size to obtain hiding capacity.

6. Insert marker FFFE and embed appended data and secret data following it in JPEG header.

The hiding procedure is shown in Fig.4,

[insert Figure 4]

## 3.3. Extract secret data and restore marked image

Extraction of secret data is just opposite. It takes following steps:

1. Extract bitstreams with marker FFFE in header, secret range is obtained according to 4 bytes following FFFE.

2. Extract terminate point and special block numbers in the next 3 bytes then secret data and appended data can be divided.

3. Extract secret data and convert it to bitstreams.

Recovery of marked image is similar with extraction that step1 and step2 are the same, and the rest is DCT coefficients decompression. Decompress JPEG file into 8×8 DCT blocks except special blocks which are recorded in file header, coefficients below terminate point are decoded in JPEG codes while coefficients above are decoded in the Rxy codes, and then decode DCT coefficients to images further.

Extraction and recovery procedures are shown in Fig.5,

[insert Figure 5]

## 4. Results

There are 64×64 blocks in a 512×512 image and we take 2 as threshold in a large number of experiments.

We take eight images in the standard library [35], and compress them in Quality Factor (QF) 90,80,70,60,50,40,30,20,10 as original host images. Secret data is random bitstreams.

One measure in this paper is embedding capacity (EC) which is secret data size in concealment, and another is Rate that refers to proportion of embedding capacity in host file.

The platform in experiment is win7 64 bits system, software is matlab2013b, and CPU is e3v3 Xeon processor with four cores in 3.3GHz.

Experimental results are shown in Table 8,

[insert Table 8]

In Table 8, compared with papers [30][33], this approach outperforms them in EC.

In [30], an alternative data hiding algorithm was proposed that in each category the maximum hiding capacity could be obtained. In category $C_i$ with $i$ bits, mapping efficiency was improved greatly that an used VLC was mapped to a specific unused VLC. It was a perfect innovation that different combination of unused VLCs could denote secret data. Although EC was large to some extent, it was still insufficient in individual categories. Then a VLC reordering algorithm in paper [27] was proposed again that shuffled VLCs in $C_i$ could denote secret data. The maximum capacity in category $C_i$ was obtained between them that total capacity was the sum in each category. The redundancy in Huffman coding was well eliminated but data relativity in quantified DCT domain was not considered that more space could be reserved.

In [33], another LDH scheme was that secret data was embedded in encrypted JPEG bitstreams. The method had a good performance for privacy preserving that concealment was made after original JPEG bitstreams encryption, and data hider did not have a knowledgement of image contents. The 2-bit string which represented a value in {-3, -2, 2, 3} was selected and lossless recompressed, then the recompressed bitstreams was encrypted with secret data embedded in the end. The structure of modified bitstreams was in JPEG format but with block numbers increasing, rate in {0, 1} also approached to 0.5 and biases in {0, 1} declined to almost 0 that EC was small.

In this paper, we select coefficients {-2, -1, 1, 2} with 2 bits which are mostly in high frequencies so that little deviation will occur. In these non-zero values, there are often some zero coefficients interval that extra bits are needed to encode them. We recode these coefficients then redundancy is eliminated greatly. In this paper, correlation in values {-2, -1, 1, 2} is made full use of and capacity is larger than state-of-the-art works.

Rate is also shown in Table 9,

[insert Table 9]

It further shows that when QF is low, quantization step is large, then non-zero coefficients are small with a lot of zeroes interval that terminate point is quite low. Advantage in this paper is not significant that relative proportion to original data is large but EC is opposite because host file is small. When QF increases, absolute values become larger, although terminate point is higher, compressed file expands, then Rate decreases a little and EC increases greatly. When QF increases further, there is still some improvement in hiding capacity generally, but in some extreme images, absolute values with high frequencies are often larger than 2, then terminate point will become very high and even exceed 64 that EC may decrease to some extent. So when QF is 70, both EC and Rate achieve good results.

## 5. Discussion

We are achieving a perfect balance of encoding efficiency in high frequencies and low frequencies. The JPEG Huffman table gives a priority in low frequencies, so we propose a segment coding scheme that coefficients below terminate point are encoded in the standard codes and coefficients above are encoded in the proposed Rxy codes. The Rxy encoding scheme is to keep non-zero range that it is only needed to recode zeroes number, then encoding length is shorten because information entropy which refers to non-zero size reduces. Thus spare space is reserved and we insert appended data and secret data in JPEG header following comment marker FFFE that concealment is lossless and hiding efficiency is improved with no file expansion.

## Acknowledgements

This work was supported by Natural Science Foundation of China (Grant 61372175).

# Author introduction

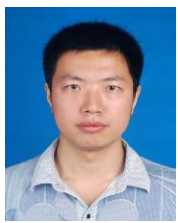

**Mingming ZHANG** received the B.S. degree in simulation engineering from National University of Defense Technology, Changsha, P.R.China in 2011. He received the M.S. degree in communication engineering from China Academy of Space Technology, Xi'an, P.R.China in 2015. He is currently working toward the Ph.D. degree in aircraft designing at China Academy of Space Technology, Xi'an, P.R.China. His research interests include data hiding and image compression.

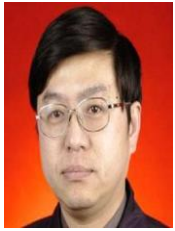

**Quan ZHOU** received the B.S., M.S. and Ph.D. degree in communication engineering from XIDIAN University, Xi'an, P.R.China in 1986, 1989 and 1992 respectively. He is currently working as a full professor in National Key Laboratory of Science and Technology on Space Microwave at China Academy of Space Technology, Xi'an, P.R.China. His research interests include digital image processing, data hiding and image compression.

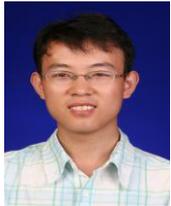

**Yanlang HU** received the B.S. degree in electronic and information engineering from Hohai University, Nanjing, P.R.China in 2005. He received the M.S. degree in communication engineering from China Academy of Space Technology, Xi'an, P.R.China, in 2008. He is currently working as a senior engineer in National Key Laboratory of Science and Technology on Space Microwave at China Academy of Space Technology, Xi'an, P.R.China. His research interests include computer vision and data transmission.

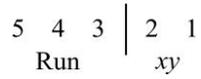

Fig.1. Rxy encoding

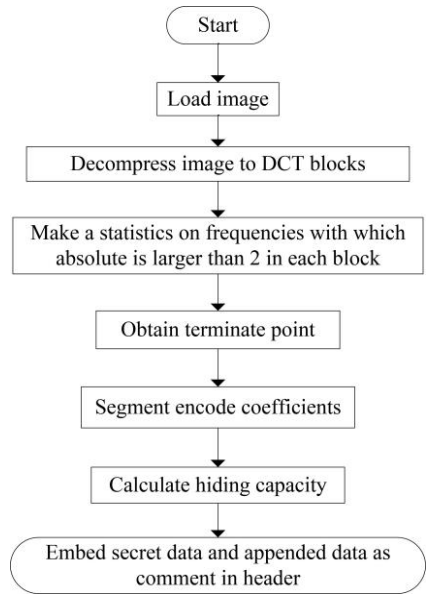

Fig.2. Data embedding flowchart

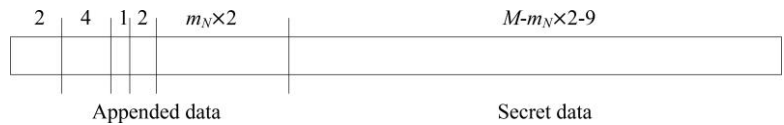

Fig.3. Additional data constitute

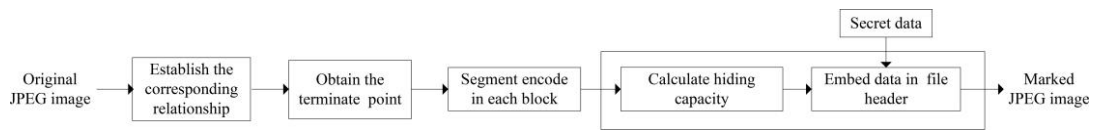

Fig.4. Concealment procedure sketch

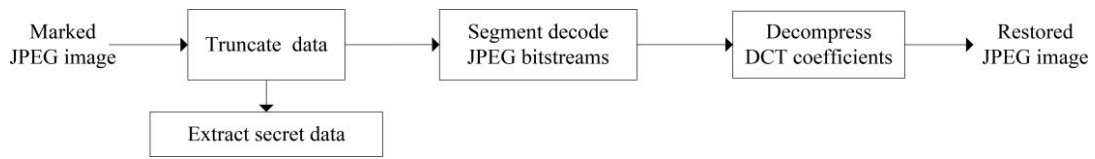

Fig.5. Extraction and recovery procedure sketch

Table 1 Brightness AC coefficients

| Run/Size | code length | code word | appended length | sum length | rate |
|---|---|---|---|---|---|
| 0/0 (EOB) | 0 | 1010 | 0 | 4 | 0 |
| 0/1 | 2 | 00 | 1 | 3 | 0.5 |
| 0/2 | 2 | 01 | 2 | 4 | 1 |
| 1/1 | 4 | 1100 | 1 | 5 | 0.25 |
| 1/2 | 5 | 11011 | 2 | 7 | 0.4 |
| 2/1 | 5 | 11100 | 1 | 6 | 0.2 |
| 2/2 | 8 | 11111001 | 2 | 10 | 0.25 |
| 3/1 | 6 | 111010 | 1 | 7 | 0.17 |
| 3/2 | 9 | 111110111 | 2 | 11 | 0.22 |
| 4/1 | 6 | 111011 | 1 | 7 | 0.17 |
| 4/2 | 10 | 1111111000 | 2 | 12 | 0.2 |
| 5/1 | 7 | 1111010 | 1 | 8 | 0.14 |
| 5/2 | 11 | 11111110111 | 2 | 13 | 0.18 |
| …… | …… | …… | …… | …… | …… |

Table 2 *Run* encoding

| Run | zeroes number |
|---|---|
| 001 | 0 |
| 010 | 1 |
| 011 | 2 |
| 100 | 3 |
| 101 | 4 |
| 110 | 5 |
| 111 | 6 |

Table 3 Corresponding *xy* codes

| xy | value |
|---|---|
| 00 | -2 |
| 01 | -1 |
| 10 | 1 |
| 11 | 2 |

Table 4 Joint coding framework

| M | Run | Rxy |
|---|---|---|
| 0 | 001 | 001*xy* |
| 1 | 010 | 010*xy* |
| 2 | 011 | 011*xy* |
| 3 | 100 | 100*xy* |
| 4 | 101 | 101*xy* |
| 5 | 110 | 110*xy* |
| 6 | 111 | 111001*xy* |

Table 5 Rxy codes

| Run/Size | code length | code word | sum length | rate |
|---|---|---|---|---|
| 0/0 (EOB) | 3 | 000 | 3 | 0 |
| 0/xy | 3 | 001 | 5 | 0.67 |
| 1/xy | 3 | 010 | 5 | 0.67 |
| 2/xy | 3 | 011 | 5 | 0.67 |
| 3/xy | 3 | 100 | 5 | 0.67 |
| 4/xy | 3 | 101 | 5 | 0.67 |
| 5/xy | 3 | 110 | 5 | 0.67 |
| 6/xy | 3+3 | 111001 | 8 | 0.33 |
| 7/xy | 3+3 | 111010 | 8 | 0.33 |
| 8/xy | 3+3 | 111011 | 8 | 0.33 |
| 9/xy | 3+3 | 111100 | 8 | 0.33 |
| 10/xy | 3+3 | 111101 | 8 | 0.33 |
| 11/xy | 3+3 | 111110 | 8 | 0.33 |
| 12/xy | 3+3+3 | 111111001 | 11 | 0.22 |
| 13/xy | 3+3+3 | 111111010 | 11 | 0.22 |
| 14/xy | 3+3+3 | 111111011 | 11 | 0.22 |
| 15/xy | 3+3+3 | 111111100 | 11 | 0.22 |
| M/xy | 3×chu+3 | | 3×chu+5 | 2/(3×chu+5) |
| …… | …… | …… | …… | …… |

Table 6 An example of 8×8 DCT block

| 12 | 17 | 10 | -5 | 0 | -2 | 0 | 0 |
|---|---|---|---|---|---|---|---|
| 15 | 8 | 0 | 0 | 0 | 0 | 0 | 0 |
| -9 | 0 | 0 | 0 | 0 | -1 | 0 | 0 |
| 3 | 4 | 0 | 0 | 0 | 0 | 0 | 0 |
| 0 | 0 | 0 | 0 | 0 | 0 | 0 | 0 |
| 1 | 0 | 0 | 0 | 0 | 0 | 0 | 0 |
| 0 | 0 | 0 | 0 | 0 | 0 | 0 | 0 |
| 0 | 0 | 0 | 0 | 0 | 0 | 0 | 0 |

Table 7 *Run/Size* in 8×8 DCT block

| 0/4 | 0/5 | 0/4 | 0/3 | 0 | 3/2 | 0 | 0 |
|---|---|---|---|---|---|---|---|
| 0/4 | 0/4 | 0 | 0 | 0 | 0 | 0 | 0 |
| 0/4 | 0 | 0 | 0 | 0 | 9/1 | 0 | 0 |
| 3/2 | 2/3 | 0 | 0 | 0 | 0 | 0 | 0 |
| 0 | 0 | 0 | 0 | 0 | 0 | 0 | 0 |
| 4/1 | 0 | 0 | 0 | 0 | 0 | 0 | 0 |
| 0 | 0 | 0 | 0 | 0 | 0 | 0 | 0 |
| 0 | 0 | 0 | 0 | 0 | 0 | 0 | 0 |

Table 8 Comparisons of EC (bits) versus QF

| Image | Method | 10 | 20 | 30 | 40 | 50 | 60 | 70 | 80 | 90 |
|---|---|---|---|---|---|---|---|---|---|---|
| Airplane | Chang[33] | 395 | 527 | 601 | 675 | 823 | 943 | 1003 | 1204 | 1275 |
|  | Qiu[30] | 1486 | 1066 | 1097 | 1215 | 1179 | 1160 | 1069 | 881 | 913 |
|  | Proposed | 2812 | 3284 | 3962 | 4630 | 5020 | 5402 | 6184 | 6312 | 6408 |
| Bridge | Chang[33] | 307 | 374 | 425 | 462 | 510 | 542 | 580 | 634 | 660 |
|  | Qiu[30] | 2473 | 1514 | 1099 | 1193 | 1269 | 926 | 1002 | 782 | 858 |
|  | Proposed | 3097 | 3220 | 3530 | 4372 | 4543 | 5462 | 5773 | 5530 | 5018 |
| Baboon | Chang[33] | 1145 | 1207 | 1213 | 1275 | 1233 | 1420 | 1499 | 1555 | 1686 |
|  | Qiu[30] | 7169 | 3767 | 1970 | 2077 | 1482 | 1582 | 1634 | 1033 | 939 |
|  | Proposed | 4270 | 4975 | 5467 | 6138 | 6529 | 5240 | 4639 | 3875 | 3052 |
| Boat | Chang[33] | 359 | 409 | 456 | 562 | 722 | 854 | 930 | 1032 | 1204 |
|  | Qiu[30] | 1635 | 1185 | 1242 | 1020 | 1079 | 1139 | 962 | 1249 | 2172 |
|  | Proposed | 3174 | 4029 | 4328 | 4937 | 5502 | 5720 | 5967 | 6021 | 6042 |
| Couple | Chang[33] | 385 | 445 | 541 | 620 | 747 | 854 | 921 | 1088 | 1204 |
|  | Qiu[30] | 1035 | 1165 | 1324 | 1054 | 1204 | 1487 | 1275 | 1053 | 1145 |
|  | Proposed | 3398 | 4667 | 5531 | 6382 | 6667 | 6712 | 7063 | 7245 | 7309 |
| Lena | Chang[33] | 298 | 332 | 452 | 489 | 528 | 596 | 675 | 798 | 905 |
|  | Qiu[30] | 1256 | 1004 | 913 | 938 | 999 | 767 | 760 | 695 | 842 |
|  | Proposed | 2067 | 3216 | 3843 | 4861 | 5342 | 5703 | 6857 | 7049 | 7260 |
| Peppers | Chang[33] | 245 | 291 | 302 | 385 | 450 | 568 | 820 | 960 | 1085 |
|  | Qiu[30] | 1263 | 989 | 1078 | 923 | 928 | 964 | 774 | 695 | 1775 |
|  | Proposed | 1713 | 2153 | 3368 | 3750 | 4682 | 5233 | 5430 | 5580 | 5549 |
| Elaine | Chang[33] | 341 | 401 | 541 | 584 | 624 | 765 | 801 | 968 | 1041 |
|  | Qiu[30] | 1036 | 944 | 842 | 933 | 974 | 763 | 781 | 997 | 1938 |
|  | Proposed | 1811 | 2338 | 2620 | 3665 | 4020 | 4310 | 4531 | 4752 | 4658 |

Table 9 Rate (%) in concealment versus QF

| Image | 10 | 20 | 30 | 40 | 50 | 60 | 70 | 80 | 90 |
|---|---|---|---|---|---|---|---|---|---|
| Airplane | 3.98 | 3.12 | 2.90 | 2.96 | 2.66 | 2.69 | 2.24 | 1.85 | 1.47 |
| Bridge | 2.89 | 1.81 | 1.45 | 1.58 | 1.32 | 1.50 | 1.10 | 0.89 | 0.62 |
| Baboon | 3.56 | 2.53 | 2.03 | 1.99 | 1.72 | 1.30 | 0.81 | 0.57 | 0.41 |
| Boat | 4.16 | 3.38 | 2.69 | 2.70 | 2.43 | 2.39 | 1.68 | 1.36 | 1.06 |
| Couple | 4.20 | 3.65 | 3.19 | 3.25 | 2.74 | 2.60 | 1.97 | 1.66 | 1.30 |
| Lena | 3.24 | 3.37 | 3.11 | 3.31 | 3.06 | 3.03 | 2.58 | 2.08 | 1.68 |
| Peppers | 2.66 | 2.29 | 2.73 | 2.60 | 2.63 | 2.72 | 1.93 | 1.56 | 1.21 |
| Elaine | 3.19 | 2.70 | 2.26 | 2.53 | 2.29 | 2.07 | 1.74 | 1.34 | 0.77 |